\documentclass[11pt,a4paper]{article}
\usepackage[hyperref]{emnlp2018}
\usepackage{times}
\usepackage{latexsym}
\usepackage{graphicx}
\usepackage{amsmath}

\usepackage{url}
\DeclareMathOperator*{\argmax}{arg\,max}

\aclfinalcopy 

\title{Crowdsourcing Semantic Label Propagation in Relation Classification}

\author{Anca Dumitrache \\
  Vrije Universiteit Amsterdam \\
  IBM CAS Benelux \\
  {\tt anca.dmtrch@gmail.com} \\\And
  Lora Aroyo \\
  Vrije Universiteit Amsterdam \\
  {\tt l.m.aroyo@gmail.com} \\\And
  Chris Welty \\
  Google Research, New York \\
  {\tt cawelty@gmail.com} \\}

\date{}

\begin{document}
\maketitle
\begin{abstract}
Distant supervision is a popular method for performing relation extraction from text that is known to produce noisy labels.  Most progress in relation extraction and classification has been made with crowdsourced corrections to distant-supervised labels, and there is evidence that indicates still more would be better.  In this paper, we explore the problem of propagating human annotation signals gathered for open-domain relation classification through the CrowdTruth methodology for crowdsourcing, that captures ambiguity in annotations by measuring inter-annotator disagreement. Our approach propagates annotations to sentences that are similar in a low dimensional embedding space, expanding the number of labels by two orders of magnitude.  Our experiments show significant improvement in a sentence-level multi-class relation classifier.
\end{abstract}

\section{Introduction}

Distant supervision (DS)~\cite{mintz2009distant} is a popular method for performing relation extraction from text. It is based on the assumption that, when a knowledge-base contains a relation between a pair of terms, then any sentence that contains that pair is likely to express the relation. This approach can generate false positives, as not every mention of a term pair in a sentence means a relation is also expressed~\cite{DBLP:conf/ijcai/FengGQLL17}. 

Recent results~\cite{angeli2014combining,liu2016effective} have shown strong evidence that the community needs more annotated data to improve the quality of DS data. This work explores the possibility of automatically expanding smaller human-annotated datasets to DS scale. \citet{sterckx2016knowledge} proposed a method to correct labels of sentence dependency paths by using expert annotators, and then propagating the corrected labels to a corpus of DS sentences by calculating the similarity between the labeled and unlabeled sentences in the embedding space of their dependency paths.

In this paper, we adapt and simplify semantic label propagation to propagate labels without computing dependency paths, and using the crowd instead of experts, which is more scalable. Our simplified algorithm propagates crowdsourced annotations from a small sample of sentences to a large DS corpus. To evaluate our approach, we perform an experiment in open domain relation classification in the English-language, using a corpus of sentences~\cite{dumitrache2017false} whose labels have been collected using the CrowdTruth method~\cite{aroyo2014threesides}.  


\section{Related Work}

\begin{figure*}[t!]
\centering
\caption{Fragment of the crowdsourcing task template.}
\label{fig:template}
\includegraphics[width=\textwidth]{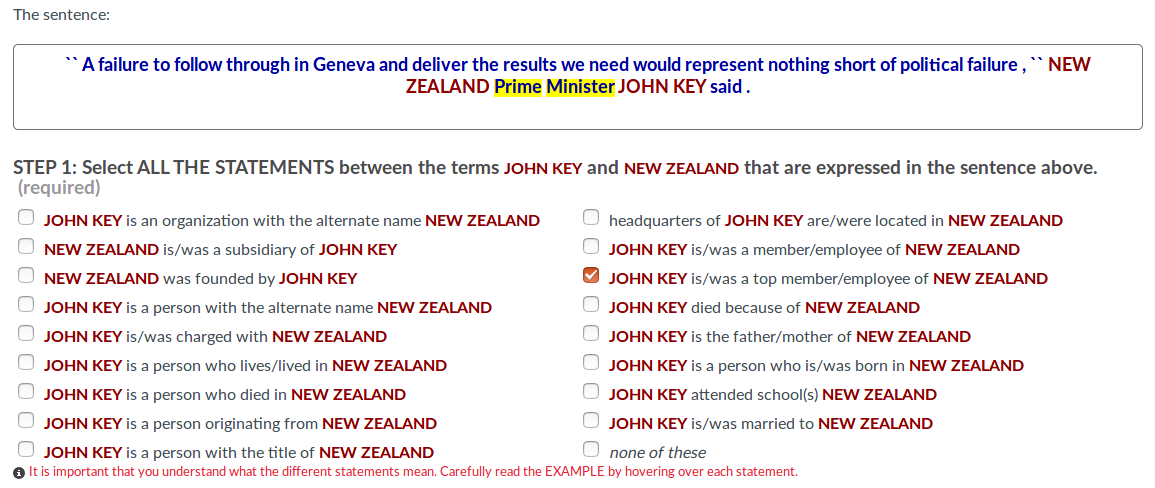}
\end{figure*}

There exist several efforts to correct DS with the help of crowdsourcing. \citet{angeli2014combining} present an active learning approach to select the most useful sentences that need human re-labeling using a query by committee. \citet{zhang2012big} show that labeled data has a statistically significant, but relatively low impact on improving the quality of DS training data, while increasing the size of the DS corpus has a more significant impact. In contrast, ~\citet{liu2016effective} prove that a corpus of labeled sentences from a pool of highly qualified workers can significantly improve DS quality. All of these methods employ large annotated corpora of 10,000 to 20,000 sentences. In our experiment, we show that a comparatively smaller corpus of 2,050 sentences is enough to correct DS errors through semantic label propagation.

\citet{levy2017zero} have shown that a small crowdsourced dataset of questions about relations can be exploited to perform zero-shot learning for relation extraction. \citet{pershina2014infusion} use a small dataset of hand-labeled data to generate relation-specific guidelines that are used as additional features in the relation extraction. The label propagation method was introduced by \citet{xiaojin2002learning}, while \citet{Chen:2006:REU:1220175.1220192} first applied it to correct DS, by calculating similarity between labeled and unlabeled examples an extensive list of features, including part-of-speech tags and target entity types. In contrast, our approach calculates similarity between examples in the word2vec~\cite{mikolov2013distributed} feature space, which it then uses to correct the labels of training sentences. This makes it easy to reuse by the state-of-the-art in both relation classification and relation extraction -- convolutional~\cite{ji2017distant} and recurrent neural network methods~\cite{zhou2016attention} that do not use extensive feature sets. To evaluate our approach, we used a simple convolutional neural network to perform relation classification in sentences~\cite{nguyen2015relation}.


\section{Experimental Setup}

\subsection{Annotated Data}



The labeled data used in our experiments consists of 4,100 sentences: 2,050 sentences from the CrowdTruth corpus~\cite{dumitrache2017false}, which we have augmented by another 2,050 sentences picked at random from the corpus of~\citet{angeli2014combining}. The resulting corpus contains sentences for 16 popular relations from the open domain, as shown in in Figure~\ref{fig:template},\footnote{The $alternate\_names$ relation appears twice in the list, once referring to alternate names of persons, and the other referring to organizations.} as well as candidate term pairs and DS seed relations for each sentence. As some relations are more general than others, the relation frequency in the corpus is slightly unequal -- e.g. $places\_of\_residence$ is more likely to be in a sentence when $place\_of\_birth$ and $place\_of\_death$ occur, but not the opposite.

The crowdsourcing task (Figure~\ref{fig:template}) was designed in our previous work~\cite{dumitrache2017false}. We asked workers to read the given sentence where the candidate term pair is highlighted, and then pick between the 16 relations or {\it none of the above}, if none of the presented relations apply. The task was multiple choice and run on the Figure Eight\footnote{\url{https://www.figure-eight.com/}} and Amazon Mechanical Turk\footnote{\url{https://www.mturk.com/}} crowdsourcing platforms. Each sentence was judged by 15 workers, and each worker was paid \$0.05 per sentence. 

Crowdsourcing annotations are aggregated usually by measuring the consensus of the workers (e.g. using majority vote). This is based on the assumption that a single right annotation exists for each example. In the problem of relation classification, the notion of a single truth is reflected in the fact that a majority of proposed solutions treat relations as mutually exclusive, and the objective of the classification task is usually to find the best relation for a given sentence and term pair. In contrast, the CrowdTruth methodology proposes that crowd annotations are inherently diverse~\cite{aroyo2015truth}, due to a variety of factors such as the ambiguity that is inherent in natural language. We use a comparatively large number of workers per sentences (15) in order to collect inter-annotator disagreement, which results in a more fine-grained ground truth that separates between clear and ambiguous expressions of relations. This is achieved by labeling examples with the inter-annotator agreement on a continuous scale, as opposed to using binary labels.


To aggregate the results of the crowd, we use CrowdTruth metrics\footnote{\url{https://github.com/CrowdTruth/CrowdTruth-core}}~\cite{dumitrache2018crowdtruth} to capture and interpret inter-annotator disagreement as quality metrics for the workers, sentences, and relations in the corpus. The annotations of one worker over one sentence are encoded as a binary worker vector with 17 components, one for each relation and including $none$. The quality metrics for the workers, sentences and relations, are based on average cosine similarity over the worker vectors --  e.g. the quality of a worker $w$ is given by the average cosine similarity between the worker vector of $w$ and the vectors of all other workers that annotated the same sentences. 
These metrics are mutually dependent (e.g. the sentence quality is weighted by the relation quality and worker quality), the intuition being that low quality workers should not count as much in determining sentence quality, and ambiguous sentences should have less of an impact in determining worker quality, etc. 

We reused these scores in our experiment, focusing on the {\bf sentence-relation score ($srs$)}, representing the degree to which a relation is expressed in the sentence. It is the ratio of workers that picked the relation to all the workers that read the sentence, weighted by the worker and relation quality.  A higher $srs$ should indicate that the relation is more clearly expressed in a sentence.

\subsection{Propagating Annotations}

Inspired by the semantic label propagation method~\cite{sterckx2016knowledge}, we propagate the vectors of $srs$ scores on each crowd annotated sentence to a much larger set of distant supervised (DS) sentences (see datasets description in Section~\ref{sec:train}), scaling the vectors linearly by the distance in low dimensional word2vec vector space~\cite{mikolov2013distributed}.  One of the reasons we chose the CrowdTruth set for this experiment is that the annotation vectors give us a score \emph{for each relation} to propagate to the DS sentences, which have only one binary label.

 Similarly to~\citet{sultan2015dls}, we calculate the vector representation of a sentence as the average over its word vectors, and like \citet{sterckx2016knowledge} we get the similarity between sentences using cosine similarity. Additionally, we restrict the sentence representation to only contain the words between the term pair, in order to reduce the vector space to the one that is most likely to express the relations. For each sentence $s$ in the DS dataset, we find the sentence $l'$ from the crowd annotated set that is most similar to $s$: $ l' = \argmax\limits_{l \in Crowd}cos\_sim(l, s). $ The score for relation $r$ of sentence $s$ is calculated as the weighted average between the $srs(l', r)$ and the original DS annotation, weighted by the cosine similarity to $s$ ($cos\_sim(s,s) = 1$ for the DS term, and $cos\_sim(s, l')$ for the $srs$ term):
\begin{equation} \label{eq:ds_w2v}
DS^{*}(s, r) =  \dfrac{DS(s, r) + cos\_sim(s, l') \cdot srs(l', r)}{1 + cos\_sim(s, l')}
\end{equation}
\noindent where $DS(s, r) \in \{0,1\}$ is the original DS annotation for the relation $r$ on sentence $s$.

\subsection{Training the Model}
\label{sec:train}

The crowdsourced data is split evenly into a dev and a test set of 2,050 sentences each chosen at random. In addition, we used a training set of 235,000 sentences annotated by DS from freebase relations, used in~\citet{riedel2013relation}.

The relation classification model employed is based on~\citet{nguyen2015relation}, who implement a convolutional neural network with four main layers: an embedding layer for the words in the sentence and the position of the candidate term pair in the sentence, a convolutional layer with a sliding window of variable length of 2 to 5 words that recognizes n-grams, a pooling layer that determines the most relevant features, and a softmax layer to perform classification.

We have adapted this model to be both multi-class and multi-label -- we use a sigmoid cross-entropy loss function instead of softmax cross-entropy, and the final layer is normalized with the sigmoid function instead of softmax -- in order to make it possible for more than one relation to hold between two terms in one sentence. The loss function is computed using continuous labels instead of binary positive/negative labels, in order to accommodate the use of the $srs$ in training. The features of the model are the word2vec embeddings of the words in the sentences, together with the position embeddings of the two terms that express the relation. The word embeddings are initialized with 300-dimensional word2vec vectors pre-trained on the Google News corpus\footnote{\url{https://code.google.com/archive/p/word2vec/}}. Both the position and word embeddings are nonstatic and become optimized during training of the model. The model is trained for 25,000 iterations, after the point of stabilization for the train loss.  The values of the other hyper-parameters are the same as those reported by \citet{nguyen2015relation}. The model was implemented in Tensorflow~\cite{abadi2016tensorflow}, and trained in a distributed manner on the DAS-5 cluster~\cite{bal2016medium}.

\begin{figure}[t!]
\caption{Precision / Recall curve, calculated for each sentence-relation pair.}
\label{fig:pr}
\includegraphics[width=0.5\textwidth]{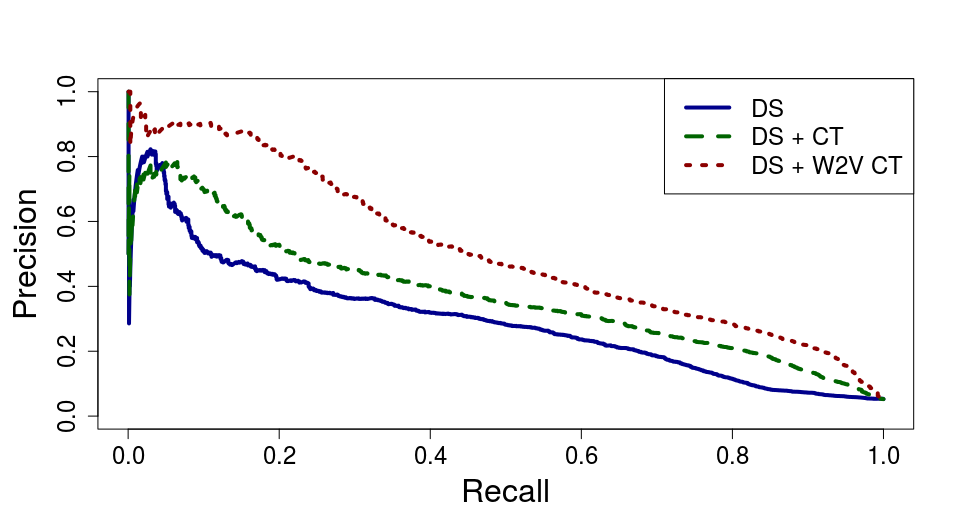}
\end{figure}

For our experiment, we split the crowd data into a dev and a test set of equal size, and compared the performance of the model on the held-out test set when trained by the following datasets:

\begin{enumerate}
\item {\bf DS:} The 235,000 sentences annotated by DS.
\item {\bf DS + CT:} The 2,050 crowd dev annotated sentences added directly to the DS dataset.
\item {\bf DS + W2V CT:} The DS$^{*}$ dataset (Eq.~\ref{eq:ds_w2v}), with relation scores propagated over the 2,050 crowd dev sentences.
\end{enumerate}

\section{Results and Discussion}

\begin{figure}[t!]
\caption{Distribution of sentence-level cosine similarity with test set values.}
\label{fig:cos_sim}
\includegraphics[width=0.5\textwidth]{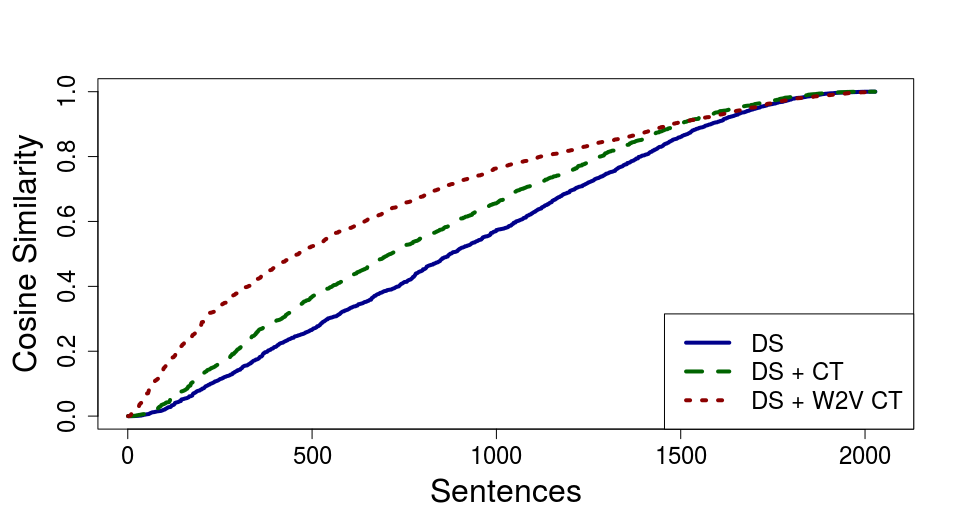}
\end{figure}

To evaluate the performance of the models, we calculate the micro precision and recall (Figure~\ref{fig:pr}), as well as the cosine similarity per sentence with the test set (Figure~\ref{fig:cos_sim}).  In order to calculate the precision and recall, a threshold of 0.5 was set in the $srs$, and each sentence-relation pair was labeled either as positive or negative. However, for calculating the cosine similarity, the $srs$ was used without change, in order to better reflect the degree of agreement the crowd had over annotating each example.  We observe that {\bf DS + W2V CT}, with a precision/recall $AUC = 0.512$, significantly outperforms {\bf DS} (P/R $AUC = 0.294$). {\bf DS + CT} (P/R $AUC = 0.372$) also does slightly better than {\bf DS}, but not enough to compete with the semantic label propagation method. The cosine similarity result (Figure~\ref{fig:cos_sim}) shows that {\bf DS + W2V CT} also produces model predictions that are closer to the different agreement levels of the crowd. Take advantage of the agreement scores in the CrowdTruth corpus, the cosine similarity evaluation allows us to assess relation confidence scores on a continuous scale. The crowdsourcing results and model predictions are available online.\footnote{\url{https://github.com/CrowdTruth/Open-Domain-Relation-Extraction}}

\begin{figure}[h!]
\centering
\caption{DS false positive ratio in combined crowd dev and test sets.}
\label{fig:fp}
\includegraphics[width=0.5\textwidth]{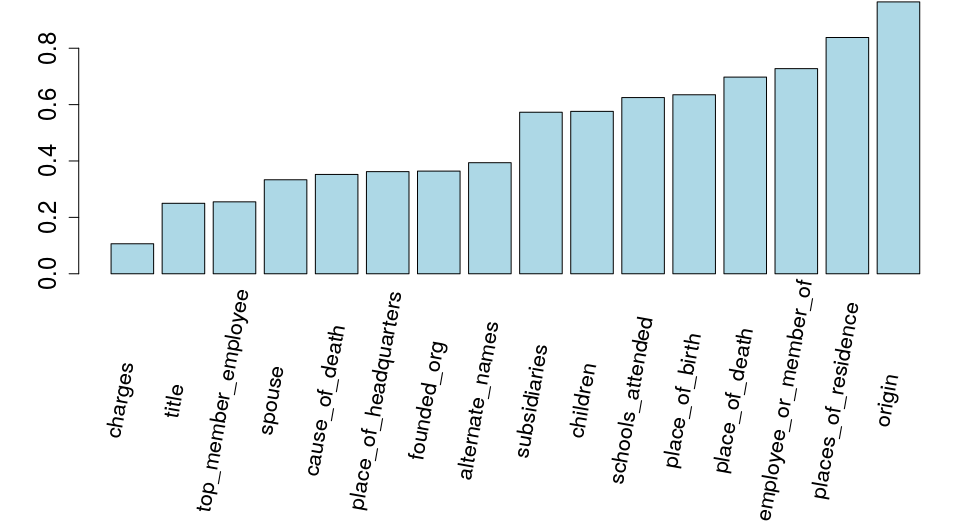}
\end{figure}

One reason for which the semantic label propagation method works better than simply adding the correctly labeled sentences to the train set is the high rate of incorrectly labeled examples in the DS training data. Figure~\ref{fig:fp} shows that some relations, such as $origin$ and $places\_of\_residence$, have a ratio of over 0.8 false positive sentences, meaning that a vast majority of training examples are incorrectly labeled.  The success of the {\bf DS + W2V CT} comes in part because the method relabels all sentences in DS. Adding correctly labeled sentences to the train set would require a significantly larger corpus in order to correct the high false positive rate, but semantic label propagation only requires a small corpus (two orders of magnitude smaller than the train set) to achieve significant improvements.

\section{Conclusion and Future Work}

This paper explores the problem of propagating human annotation signals in distant supervision data for open-domain relation classification.  Our approach propagates human annotations to sentences that are similar in a low dimensional embedding space, using a small crowdsourced dataset of 2,050 sentences to correct training data labeled with distant supervision.  We present experimental results from training a relation classifier, where our method shows significant improvement over the DS baseline, as well as just adding the labeled examples to the train set.

Unlike \citet{sterckx2016knowledge} who employ experts to label the dependency path representation of sentences, our method uses the general crowd to annotate the actual sentence text, and is thus easier to scale and not dependent on methods for extracting dependency paths, so it can be more easily adapted to other languages and domains.  Also, since the semantic label propagation is applied to the data before training is completed, this method can easily be reused to correct train data for any model, regardless of the features used in learning.  In our future work, we plan to use this method to correct training data for state-of-the-art models in relation classification, but also relation extraction and knowledge-base population.

We also plan to explore different ways of collecting and aggregating data from the crowd. CrowdTruth~\cite{dumitrache2017false} proposes capturing ambiguity through inter-annotator disagreement, which necessitates multiple annotators per sentence, while \citet{liu2016effective} propose increasing the number of labeled examples added to the training set by using one high quality worker per sentence. We will compare the two methods to determine whether quality or quantity of data are more useful for semantic label propagation. To achieve this, we will investigate whether disagreement-based metrics such as sentence and relation quality can also be propagated through the training data.


\bibliography{sources}
\bibliographystyle{acl_natbib_nourl}

\end{document}